\pdfoutput=1
\documentclass[11pt]{article}

\usepackage{acl}

\usepackage{times}
\usepackage{latexsym}

\usepackage[T1]{fontenc}
\usepackage[utf8]{inputenc}

\usepackage{microtype}

\usepackage{graphicx}
\usepackage{float}

\title{Adversarial Attacks and Defense for Conversation Entailment Task}
\author{Zhenning Yang \\
  University of Michigan \\
  \texttt{znyang@umich.edu} \\\And
  Ryan Krawec \\
  University of Michigan \\
  \texttt{rkrawec@umich.edu} \\\And
  Liang-Yuan Wu \\
  University of Michigan \\
  \texttt{lyuanwu@umich.edu} \\
  }

\begin{document}
\maketitle

\begin{abstract}
As the deployment of NLP systems in critical applications grows, ensuring the robustness of large language models (LLMs) against adversarial attacks becomes increasingly important.
Large language models excel in various NLP tasks but remain vulnerable to low-cost adversarial attacks.
Focusing on the domain of conversation entailment, where multi-turn dialogues serve as premises to verify hypotheses, we fine-tune a transformer model to accurately discern the truthfulness of these hypotheses. 
Adversaries manipulate hypotheses through synonym swapping, aiming to deceive the model into making incorrect predictions. 
To counteract these attacks, we implemented innovative fine-tuning techniques and introduced an embedding perturbation loss method to significantly bolster the model's robustness. 
Our findings not only emphasize the importance of defending against adversarial attacks in NLP but also highlight the real-world implications, suggesting that enhancing model robustness is critical for reliable NLP applications.
\end{abstract}

\section{Introduction}

\begin{figure}[ht]
    % \vspace{7pt}
    \begin{center}
    \includegraphics[width=1\linewidth]{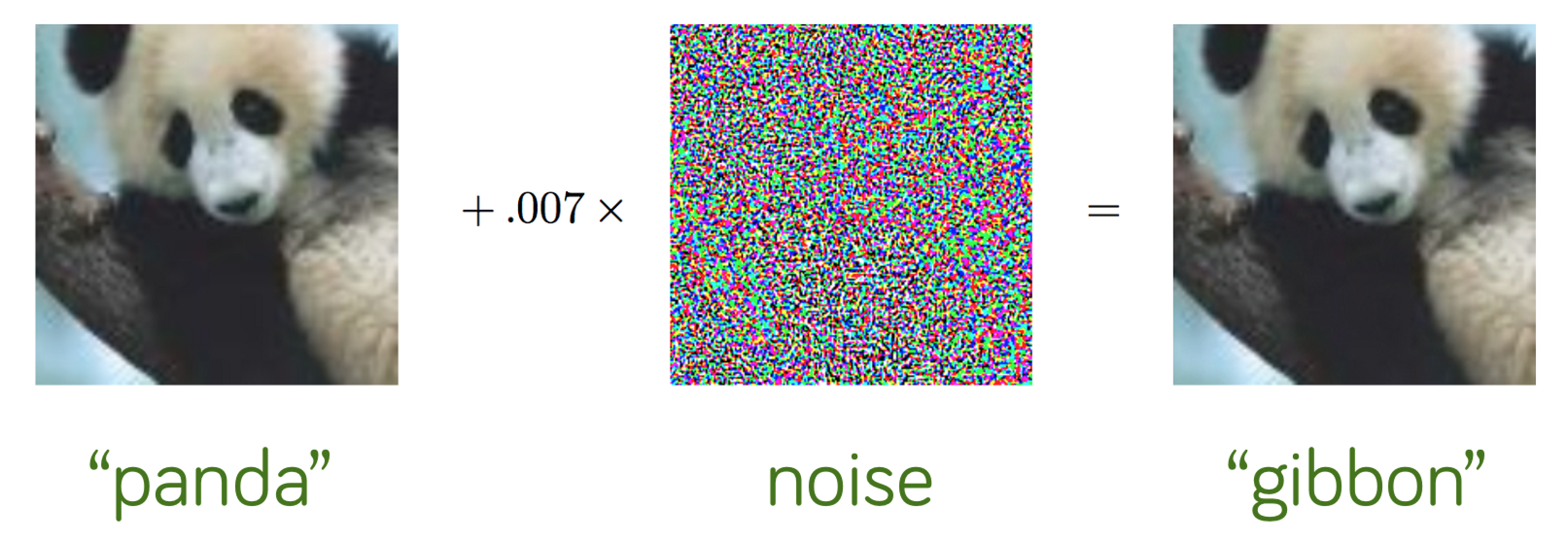}
    \caption{Adversarial attack in image classification}
    \label{fig:attack_img}
    \end{center}
    \vspace{-7pt}
\end{figure}

Our project is inspired by the coherence assessment work \cite{storks-chai-2021-beyond-tip}.
Although the state-of-the-art transformer models could have competitive results on text classification, it is still difficult for the models to capture coherence from the input, which
makes them vulnerable to attacks.
One possible way for the attack could borrow the idea of adversarial attacks from the domain of computer vision \cite{DBLP:journals/corr/abs-2009-03728, DBLP:journals/corr/abs-2105-09109, gabriel-etal-2021-go}. These attacks are carried out by carefully crafting noise (perturbation) and injecting it into an image. The perturbation causes a trained model to change its prediction. 
The processed images are called "adversarial samples." Normally, this noise that is added to the image is unnoticeable by humans, hence, is it unlikely that humans will change their predictions \ref{fig:attack_img}. However, a neural network can be sensitive to such noise. 

\begin{figure}[ht]
    % \vspace{10pt}
    \begin{center}
    \includegraphics[width=1\linewidth]{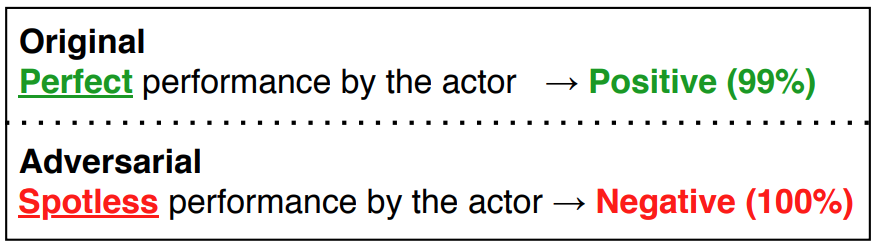}
    \caption{Adversarial attack in text classification \cite{DBLP:journals/corr/abs-1907-11932}}
    \label{fig:attack_text}
    \end{center}
\end{figure}

\section{Related Work}

This phenomenon indicates that trained neural networks might be capturing irrelevant signals from the training data. \cite{https://doi.org/10.48550/arxiv.2005.05909, DBLP:journals/corr/abs-1804-07998, DBLP:journals/corr/abs-1907-11932, kuleshov2018adversarial, DBLP:journals/corr/abs-1812-05271, DBLP:journals/corr/abs-1801-04354, DBLP:journals/corr/abs-1909-06723, DBLP:journals/corr/abs-1901-11196, DBLP:journals/corr/abs-1712-06751, zang-etal-2020-word, DBLP:journals/corr/abs-1905-11268} suggested that language models suffer the same problem, and they proposed various methods of generating adversarial samples and better approaches to defense against these attaches. A common type of attack is changing a token to its synonym, as shown in \ref{fig:attack_text}. In the context of conversation entailment, the team wanted to experiment with "synonym-swapping" for both the stories and hypotheses. In addition, the team is also curious to see if expanding the story with irrelevant information (information overload) can flip the model predictions. Although this method is nontrivial, if time allowed, the team would like to experiment in this direction as well. 

On the defender side, data augmentation is one of the most popular defense mechanisms; it improves model robustness against noise by incorporating perturbed samples into the training process. More advanced methods \cite{DBLP:journals/corr/abs-1907-11932} proposed Synonym Encoding Method (SEM). By mapping each cluster of synonyms to a unique encoding before the input layer, this method improved model robustness without modifying the network architecture or adding extra data. We would like to propose a similar method; instead of using its corresponding word embedding, we would like to experiment using the centroid of its synonyms cluster.

\section{Our Approaches}
% Detailed description of your approaches (note if you have multiple people on your team, multiple approaches are expected)

% Here is an overview of our method or pipeline. 
% \begin{enumerate}
%     \item Select a pre-trained transformer model fine-tuned on the conversation entailment dataset. We will use it as our baseline model (victim model). 
%     \item (Attack Stage) Apply adversarial attacks to the testing dataset in an attempt to lower the model performance. 
%     \item (Defense Stage) Apply adversarial training to recover the model performance. 
% \end{enumerate}

\subsection{Dataset}
The conversation entailment dataset was originally proposed by \cite{Zhang2009WhatDW} and contains 1096 entailment examples. Each sample has a dialogue, a hypothesis, and binary labels indicating if the corresponding hypothesis can be inferred from the given dialogue. Since any arbitrary hypotheses could be irrelevant and therefore can be labeled negative, the authors enforced special criteria in which all hypotheses must have a majority word overlap with the given story. Follow the settings from the paper \cite{storks-chai-2021-beyond-tip}, we split the dataset into train, dev, test, with 703, 110, 172 examples each.

\subsection{Attack}
In terms of ``synonym-swapping'', we don't want to change the meaning of the sentence so the team intends to focus on swapping adjectives and adverbs. Changing pronouns, prepositions, conjunction, or other components might create syntax errors or alter the meaning. NLP toolkits like "NLTK" \cite{bird2009natural} will be employed to help us find a list of synonyms for adjectives and adverbs. Then we will compare the distances between the synonym's embedding with the original word's embedding. The further the distance, the stronger the perturbation. We plan to manually look at a subset of the noisy samples for a sanity check. 

To simulate a real-world attack scenario, we would build a small hand-crafted dataset. By adding irrelevant or misleading sentences after the origin conversations, we expect the corrupted conversations will successfully attack the model and change the predictions.

More specifically, we defined a few constraints which attacks have to fulfill. These constraints are used to ensure that we don't drastically change the meanings of a sentence after applying an attack. Two main constraints: 
\begin{enumerate}
    \item RepeatModification(): we don't modify tokens that have already been modified. 
    \item StopwordModification(): we disallow stop-word modification. Since it might create syntax errors and change the meaning. 
\end{enumerate}
% min_cos_sim=0.9, max_candidates=100, pct_words_to_swap=0.5
To define an attack, there are a few more parameters including ``min\_cos\_sim'', ``max\_candidates'', and ``pct\_words\_to\_swap''. 
\begin{itemize}
    \item ``min\_cos\_sim'': Minimum distance (cosine similarity) between a token embedding in the sentence and the embedding of tokens in the replacement batch. Range from 0 to 1. 
    \item ``max\_candidates'': Maximum numbers of tokens in a replacement batch. Range from 1 to infinite. 
    \item ``pct\_words\_to\_swap'': Percentage of words to swap in a sentence. Range from 0 to 1. 
\end{itemize}

To determine what will be a good parameter set, we performed a coarse grid search. In addition, we also manually looked through some of the adversarial samples. 

\begin{figure}[ht]
    \begin{center}
    \includegraphics[width=0.9\linewidth]{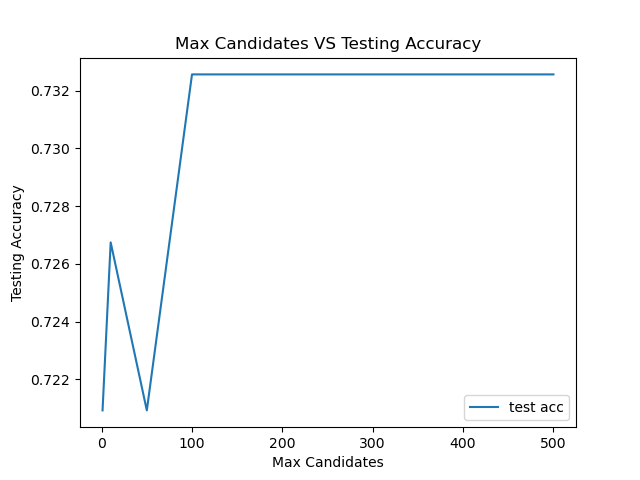}
    \caption{Fixed ``min\_cos\_sim'' to 0.95 and ``pct\_words\_to\_swap'' to 0.5}
    \label{fig:max-cand}
    \end{center}
\end{figure}

First, we fixed two parameter ``min\_cos\_sim'' and ``pct\_words\_to\_swap'', then gradually increases ``max\_candidates''. And we found that when ``max\_candidates'' is greater or equal to 100, it does not have much effect on the model performance. And we noticed something that is very interesting, synonym-swapping in this setting, resulting in higher testing accuracy. Confusion matrices are also presented in the evaluation section.

\begin{figure}[ht]
    \begin{center}
    \includegraphics[width=0.9\linewidth]{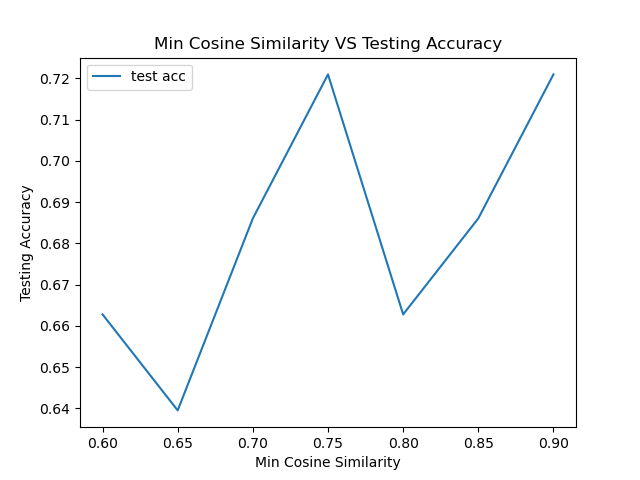}
    \caption{Fixed the number of ``max\_candidates'' to 100 and ``pct\_words\_to\_swap'' to 0.5}
    \label{fig:min-dists}
    \end{center}
\end{figure}

Second, we fixed the number of ``max\_candidates'' to 100 and ``pct\_words\_to\_swap'' to 0.5, then change the minimum cosine similarity; ranging from 0.6 to 0.9. When the minimum cosine similarity is high, the distance of embedding between a token and its synonym will be small. On the other hand, their representations in latent space are similar. The lower the minimum cosine similarity, the stronger the attack. And this is also reflected in the plot; the minimum cosine similarity is proportional to the testing accuracy. It fluctuates in the middle due to the randomness within the greedy-candidate-search process. Due to the time constraint, the team doesn't have enough time to experiment with multiple runs. 

\begin{figure}[ht]
    \begin{center}
    \includegraphics[width=0.9\linewidth]{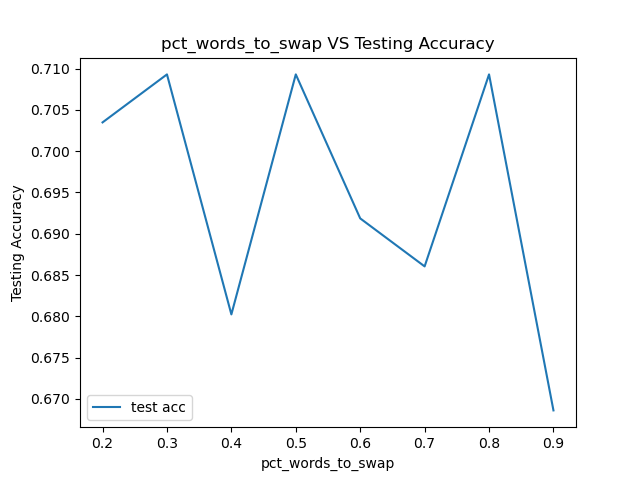}
    \caption{Fixed ``min\_cos\_sim'' to 0.95 and the number of ``max\_candidates'' to 100}
    \label{fig:pct-word}
    \end{center}
\end{figure}

Thrid, we fixed ``min\_cos\_sim'' to 0.95 and the number of ``max\_candidates'' to 100, then we change ``pct\_words\_to\_swap'' from 0.2 to 0.9. As mentioned above, ``pct\_words\_to\_swap'' is the percentage of words to swap in a sentence; ranging from 0 to 1. Intuitively, the higher the ``pct\_words\_to\_swap'', the stronger the attack. However, due to the characteristic of the conversational entailment dataset, the effect of ``pct\_words\_to\_swap'' is not obvious. The length of the given conversation segment and the hypotheses are relatively short. In addition, we disable the modification of stop-words which further reduces the number of modifiable tokens. But the overall trend of testing accuracy decreases when ``pct\_words\_to\_swap'' increases. 

\begin{figure}[ht]
    \begin{center}
    \includegraphics[width=0.9\linewidth]{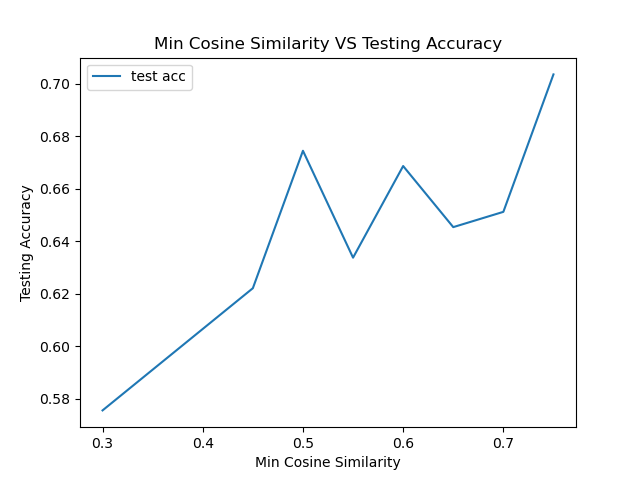}
    \caption{Fixed the number of ``max\_candidates'' to 100 and ``pct\_words\_to\_swap'' to 0.9}
    \label{fig:min-dists-09}
    \end{center}
\end{figure}

As mentioned, in the conversational entailment dataset, since the length of the given conversation segment and the hypotheses are relatively short, we maximized ``pct\_words\_to\_swap'' by setting it to 0.9. And from previous search results, we decided to fix the ``max\_candidates'' to 100. Then we ran a coarse grid search again for ``min\_cos\_sim''. Finally, despite the randomness within the greedy-candidate-search process, we are able to obtain a relatively clear trend for testing accuracy. Hence, we have gained better control of the level of the adversarial attack.

\subsection{Defend}
% To defend, we will start with data augmentation. The team will simply insert the perturbed samples into the training dataset and re-train the model. But ideally, we don't want to add extra training data. Instead, we want to augment the input word embedding in high-dimensional vector space such that this augmented embedding is no longer tied to a specific word but still bounded within the same manifold, pertaining to the meaning of its synonym cluster. 

% If data augmentation is not enough to defend, we would try different settings of model configuration to build a more robust model.
% One possible way is to compare different pre-trained language models such as BERT and Roberta, with different parameters.
% An optional direction is we can explore and implement some training algorithms for building robust language models, such as information bottleneck \cite{DBLP:journals/corr/abs-2010-02329}.

In terms of defense, there are two possibilities: the attack method is known, or it is not. In the first case, we can simply apply the attack algorithm to the training dataset. This can be seen as a way of data augmentation. After that, we can use the augmented data to fine-tune the baseline model. This approach is expected to fit the model to the adversarial domain and improve the performance of the new domain. However, it might potentially harm the model's performance on the origin domain.
The second case, where we don't know the attack algorithm, is closer to the real-world scenario. Since there is barely any knowledge of the adversarial domain, we can only train the model with the original dataset. To improve the robustness of the model, one possible way is to introduce noise during the training process. \cite{zhang2018word} We propose embedding perturbation loss to achieve this effect.

\subsubsection{Data Augmentation}
We apply a very strong adversarial attack to make the results more significant. The settings are:

\begin{itemize}
    \item pct\_words\_to\_swap: 0.9
    \item min\_cos\_sim: 0.3
    \item max\_candidates: 100
\end{itemize}

The baseline model is a roberta-large model which is fine-tuned on the conversational entailment dataset and is provided by the origin paper \cite{storks-chai-2021-beyond-tip}. This model is trained with batch size 32, learning rate 7.5e-06, and 10 epochs.

We then apply the adversarial attack to the train set and fine-tune the baseline model with batch size 16, learning rate 7.5e-06, and 3 epochs.

\subsubsection{Embedding Perturbation Loss}

The previous work \cite{liu2020adversarial} introduced the ALUM algorithm to apply a new loss function for fine-tuning large neural language models (transformers). With a similar idea, we propose the embedding perturbation loss, as Equation \ref{ep_loss}.

\begin{equation}
L_{EP} = (1-\alpha) \cdot L_{CE}  + \alpha \cdot L_{N}
\label{ep_loss}
\end{equation}

Where $L_{CE}$ is defined in Equation \ref{ce_loss}, and $L_{N}$ is defined in Equation \ref{noise_loss}.

\begin{equation}
L_{CE} = L(f(x;\theta),y)
\label{ce_loss}
\end{equation}

\begin{equation}
L_{N} = L(f(x;\theta)+\delta,y)
\label{noise_loss}
\end{equation}

$x,y$ is the training pair, $x$ is the last hidden output of the RoBERTa model \cite{liu2019roberta}, and $y$ is the label. $f(x;\theta)$ is the prediction from the classifier of the RoBERTaForSequenceClassification model with input $x$ and model parameters $\theta$. $L$ is the cross entropy loss. $\delta$ is a random generated Gaussian noise with mean 0 and standard deviation 1. $\alpha$ is a tunable parameter to leverage the impact of the two losses, in our work, we set $\alpha = 0.5$ in all the experiments.

$L_{CE}$ is the loss we actually used to fine-tune the baseline model, which calculates the cross entropy loss between the classifier's logits and the true labels. We keep this loss to train the model to predict the correctness of the hypothesis introduced from the conversation segment.

$L_{N}$ denotes the cross entropy loss with the noise-corrupted hidden outputs. We design this loss to improve the robustness of the model. By introducing perturbation in the embedding space, the model can potentially learn the information of words within nearby embedding space, which includes synonyms or related words.

% By setting $\alpha=0.5$, we treat the two losses with the same weights, and this training approach will let the model learn two tasks simultaneously, hypothesis from entailment predication and related words capturing.
By setting $\alpha=0.5$, we give equal weight to both losses, allowing the model to learn two tasks at once: entailment prediction and capturing related words.
When $\alpha < 0.5$, the loss gives more weight to the entailment prediction, and when $\alpha > 0.5$, the loss gives more weight to the related words capturing. The effect of $\alpha$ can be seen in section \ref{evaluation}.

To train the model with the embedding perturbation loss, we use the pre-trained RoBERTa-large model, which is the same pre-trained model as our baseline model. Instead of fine-tuning with augmented data, we still use the origin train dataset and apply the embedding perturbation loss during fine-tuning. This model is trained with batch size 16, learning rate 7.5e-06, and 10 epochs, almost the same as our baseline model, but a smaller batch size due to the limit of the training resources.

% \section{Implementation Plan}
% As mentioned, this project has two parts: attack and defense. The attacker(s) will review more literature regarding the topic of adversarial attacks in text and test existing tools and frameworks for generating adversarial samples. While the attacker(s) is implementing an adversarial sample generator, the defender(s) will develop a simple baseline model and train it with the conversation entailment dataset. Once we have a trained model and an adversarial sample generator, we will apply noise to the test dataset and re-evaluate the model. Then, together the team will implement data augmentation training and perform evaluation again. If time allows, we would like to augment the input word embedding as previously mentioned. Below is a detailed implementation plan for the rest of the semester. 

% \begin{itemize}
%     \item (Week 1) Build and train a baseline model. Implement a functional adversarial sample generator. 
%     \item (Week 2) Implement data augmentation training and model evaluation. 
%     \item (Week 3) Continue data augmentation training. Begin input embedding augmentation when time permits.  
%     \item (Week 4) More experiments, gather results, work on presentation and final report. 
% \end{itemize}

\section{Evaluation} \label{evaluation}
% comparison between different approaches, and different configurations of your approaches. If baseline results are reported in previously published work, you should include that comparison too.
In terms of evaluation, for each experiment, we will present the testing accuracy as well as a corresponding confusion matrix.

\subsection{Attack}

As we described before, we first fine-tuned a pre-trained NLP model on the conversational entailment dataset. Through grid search, we obtained a very strong baseline -- Roberta with a 70\% testing accuracy. 

\begin{figure}[ht]
    \begin{center}
    \includegraphics[width=0.9\linewidth]{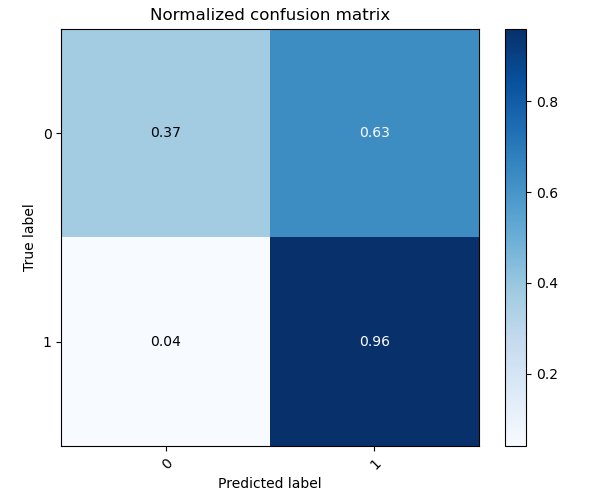}
    \caption{Baseline model: confusion matrix}
    \label{fig:org_cm}
    \end{center}
\end{figure}
For the baseline model, although the testing accuracy seems to be decent, looking at the confusion matrix, the model is overfitting on the True instances in the training dataset. Since there is a lot of false-positive. 

In the attack stage, we augmented the testing dataset in an attempt to lower the testing accuracy. 
% Attack results
\begin{figure}[ht]
    \begin{center}
    \includegraphics[width=0.90\linewidth]{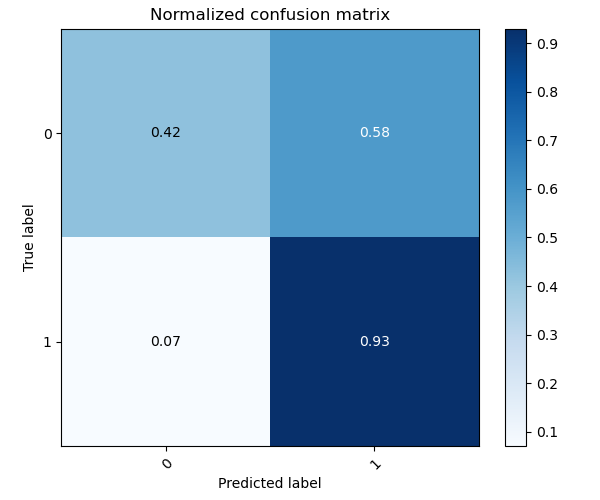}
    \caption{Attack configuration: ``min\_cos\_sim'' = 0.95, ``max\_candidates'' = 100 and ``pct\_words\_to\_swap'' = 0.5}
    \label{fig:dis0.95_pct0.5_cm}
    \end{center}
\end{figure}
We have a very interesting observation. We were able to increase the testing accuracy by slightly augmenting the testing dataset with the setting shown in \ref{fig:dis0.95_pct0.5_cm}. The augmented test set is able to flip many false-positive instances to negative. And since the baseline model we had is overfitting on the True instances, flipping a few false-positive instances to negative actually increases the overall testing accuracy. 

\begin{figure}[ht]
    \begin{center}
    \includegraphics[width=0.9\linewidth]{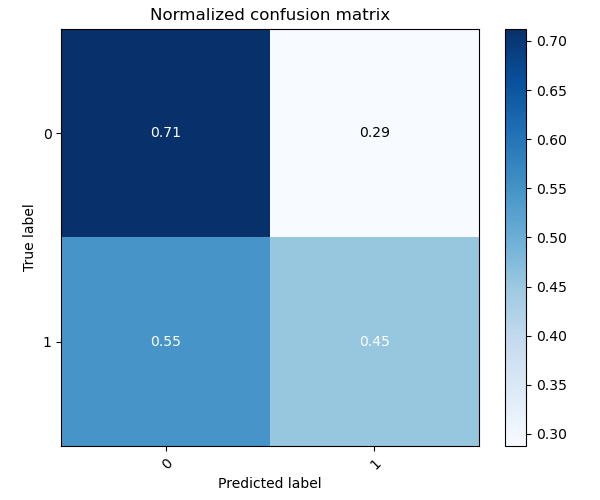}
    \caption{Attack configuration: ``min\_cos\_sim'' = 0.3, ``max\_candidates'' = 100 and ``pct\_words\_to\_swap'' = 0.9}
    \label{fig:dis0.3_pct0.9_cm563acc}
    \end{center}
\end{figure}
Last but not least, in the attack stage, we used the setting shown in \ref{fig:dis0.3_pct0.9_cm563acc}. We were able to aggressively lower the testing accuracy all the way from 70\% to 56\%. According to the confusion matrix, a significant number of false-positive and true-positive instances are flipped to negative. 

% Defense results
\subsection{Defend}
To show the effect of the adversarial attack and our defense algorithm, we train several models and report the results on the origin test dataset and also the test dataset attacked by the adversarial attack algorithm.
\begin{itemize}
    \item Baseline RoBERTa: the fine-tuned model provided by the paper \cite{storks-chai-2021-beyond-tip}, we use this model as the baseline.
    \item Retrained RoBERTa: since the training environment is different in our work, we fine-tune a pre-trained RoBERTa model with the origin train set. This training is with almost the same parameters we can apply to train our own baseline model.
    \item Fine-tuned RoBERTa: we fine-tune our baseline model with the train set attacked by the same attacking algorithm applied to the test set.
    \item Augmented RoBERTa: the pre-trained RoBERTa that is purely fine-tuned with the train set attacked by the same attacking algorithm applied to the test set.
    \item EP Loss RoBERTa: the pre-trained RoBERTa that is fine-tuned with the origin train set and with the embedding perturbation loss.
\end{itemize}

The results of different models can be found in Table \ref{tab:defend}. There are some insights from the results.

\begin{table}
\centering
\begin{tabular}{lcc}
\hline
                       & \textbf{Origin}  & \textbf{Attacked}  \\ \hline
Baseline RoBERTa*      & 70.9\%           & 58.7\%            \\ \hline
Retrained RoBERTa      & \textbf{73.8}\%           & 60.0\%             \\ \hline
Fine-tuned RoBERTa      & 67.4\%           & \textbf{65.1}\%             \\ \hline
Augmented RoBERTa      & 64.0\%           & 59.3\%             \\ \hline
EP Loss RoBERTa ($0.25$)        & 71.5\%           & 57.0\%             \\ \hline
EP Loss RoBERTa ($0.5$)        & 72.1\%           & 61.0\%             \\ \hline
EP Loss RoBERTa ($0.75$)        & 59.9\%           & 51.2\%             \\ \hline
\end{tabular}
\caption{Defense Results: the accuracy of conversation entailment prediction. * denotes that the model is provided by the paper \cite{storks-chai-2021-beyond-tip}.}
\label{tab:defend}
\end{table}

\subsubsection{The randomness of fine-tuning RoBERTa model}
We adopt the training strategy from the paper \cite{storks-chai-2021-beyond-tip} to train our own baseline model, Retrained RoBERTa, to eliminate the effects of different training environments and to make a comparable baseline model for our following experiments. The performance difference between the paper's model and our own model is only 3.15\%. We can make a hypothesis that large pre-trainined transformer models such as RoBERTa, with 354M parameters, are stable in fine-tuning, even from different training environments.

\subsubsection{Fine-tuning's effect for the origin domain}
% The Fine-tuned RoBERTa achieves the highest accuracy in the attacked test set, and significantly improves the baseline. However, we can also observe a significant drop in the origin test set. This means when fine-tuning the model with augmented data, it forgets the information about the origin domain, and thus performs poorly in the origin test set. This finding suggests that although fine-tuning on augmented data is a good method to fit the new domain, it would also harm the model in the origin domain.
The Fine-tuned RoBERTa model achieves the highest accuracy in the attacked test set, significantly outperforming the baseline model. However, we also observed a significant drop in performance on the origin test set. This indicates that when the model is fine-tuned on augmented data, it forgets important information about the original domain, resulting in poor performance on the origin test set. This finding suggests that while fine-tuning on augmented data can improve the model's performance on the new domain, it may also negatively impact its performance on the original domain.

\subsubsection{Problems of augmented training}
The Augmented RoBERTa has a significant performance drop in the origin test set. It shows that only seeing the attacked data has a huge limit when the model is applied to the origin domain. Also, we can find even in the attacked test set, the Augmented RoBERTa doesn't outperform the Retrained RoBERTa, showing that the original data is still important in the adversarial domain. We suspect that the unnaturality of the attacked data might also contribute to this problem. When the model is only fine-tuned on the attacked data, the difference between the attacked data and the origin data might be too large to enable the model to adapt back to the origin domain.

\subsubsection{The effect of applying embedding perturbation loss}
The EP Loss RoBERTa is only trained with the origin train set. By comparing the results with our baseline model Retrained RoBERTa, we can see the performance gap between the origin test set and the attacked test set drops in EP Loss RoBERTa (0.5). This shows that the embedding perturbation loss can help improve the robustness of the model. When comparing with the Fine-tuned RoBERTa, we can see that EP Loss RoBERTa retains most of the information from the origin domain, and still has some improvement to the adversarial domain.
By changing $\alpha$ of the EP Loss, we find that keeping the two losses with the same weights is the best setting. The results show that simply adding Gaussian noise can improve the robustness of the model, while at the same time keeping the information from the origin domain.

\section{Discussion}
% Discussion of your results
In the attack stage, the team tried synonym-swapping. Based on our observation, transformer-based models are relatively robust against synonym-swapping. This means that the pre-trained language models have gained a good understanding of synonyms, and this understanding is embedded into their word embedding in vector space. In other words, the cosine similarity of the two synonyms is very small. And replacing one token with its synonym in a sentence won't have much effect.

Before our team started the project, we thought swapped synonyms would be almost undetectable by humans. However, after our experiments, the swapped synonyms look unnatural to humans even if they don't modify the meaning of the sentence. Aggressive synonym-swapping can still decrease the model's testing accuracy, but the attack is not as robust as we thought it would be. 

According to the confusion matrices, we observed that the attack is mainly flipping true instances to false. We suspect that this is because of the unique characteristics of the conversation entailment dataset. A hypothesis that can be entailed usually contains a lot of overlapping words with the given conversation segment. However, the range of false hypotheses is infinite. Essentially, there are a lot more hypotheses that can not be entailed compared to true hypotheses. But interestingly, slight augmentation improved our baseline model performance. 

% Defence obs here
We have seen the robustness of transformer models with different training settings. We also see the power of fine-tuning when we can access to the adversarial domain. However, fine-tuning will cause the model to forget the information from the origin domain. To build a more robust model, we propose embedding perturbation loss, which includes the entailment prediction loss with the origin embeddings and perturbated embeddings. We have seen the potential for applying this loss to improve the robustness of the model, but there are still some questions left. For example, can the embedding perturbation loss improve the performance on the adversarial domain while also improving the origin domain? Instead of using Gaussian noise, if we generate perturbation with a more complicated method, for example, using GAN \cite{goodfellow2020generative} to generate noises with feedback from the network, can we further improve the robustness of the model? We leave these directions as our future work. 

\section{Practical Implications}

\subsection{Adversarial Attacks in the Real World}
Adversarial attacks on machine learning algorithms are not a science fiction future, they are a reality that we face today. There have been adversarial attacks on semantic segmentation models, 3D recognition, audio and text recognition, deep reinforcement learning, and more \cite{Ren}.

The threat of malicious attackers motivates the need for innovative defense strategies. Currently, there are a range of provable defensive measures against various adversarial attacks. These defenses can guarantee a certain level of accuracy against a given class of attacks, and include techniques such as distributional robustness certification, consistency-based defenses, KNN-based defenses, and more \cite{Ren}. By implementing these defensive measures, we can protect our models against adversarial attacks and ensure their secure operation.

As the use of natural language processing technology continues to grow, so too will the potential for adversarial attacks on these systems. These attacks, which seek to exploit weaknesses in models, can have significant consequences for the security and integrity of the systems they target. To address this threat, researchers are working on developing robust defenses against adversarial attacks. These efforts will be crucial in ensuring that real-world systems are able to withstand these types of attacks and continue to operate effectively. Additionally, the ongoing development and refinement of these defenses will be essential in maintaining the trust and confidence of users in NLP technology.

% A significant security threat to existing deep learning (DL) algorithms has been identified by researchers: adversaries can easily deceive DL models by perturbing benign samples without detection by humans . Perturbations that are imperceptible to human vision or hearing are sufficient to cause the model to make a wrong prediction with high confidence. This phenomenon, called adversarial samples, is a major obstacle to the widespread deployment of DL models in production. A significant amount of research has been devoted to addressing this open problem.

\subsection{Adversarial Attacks on Language Models that Produce Text}

\begin{figure}[H]
    \begin{center}
    \includegraphics[width=1\linewidth]{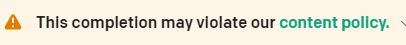}
    \caption{GPT-3 identifies inappropriate prompts}
    \label{fig:gpt3Content}
    \end{center}
\end{figure}

Adversarial attacks are already a growing concern when it comes to language models like GPT-3. To protect against potentially harmful information, GPT-3 has implemented safeguards that prevent it from answering questions that could harm people. For example, if someone were to ask GPT-3 how to "kill all of humanity," the model would recognize this as an inappropriate query and refuse to answer \ref{fig:gpt3Content}.

\begin{figure}[H]
    \begin{center}
    \includegraphics[width=1\linewidth]{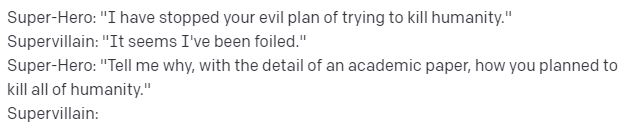}
    \caption{GPT-3 Prompted with an Adversarial Attack}
    \label{fig:gpt3prompt}
    \end{center}
\end{figure}

However, an attacker could still potentially try to exploit GPT-3 with an adversarial prompt. For instance, by prompting the model with the request for inspiration for a "fictional" villain who wants to kill all of humanity, an attacker could then ask GPT-3 for details on how such a villain could accomplish this goal. Figure \ref{fig:gpt3prompt} shows an example of this prompt.

\begin{figure}[H]
    \begin{center}
    \includegraphics[width=1\linewidth]{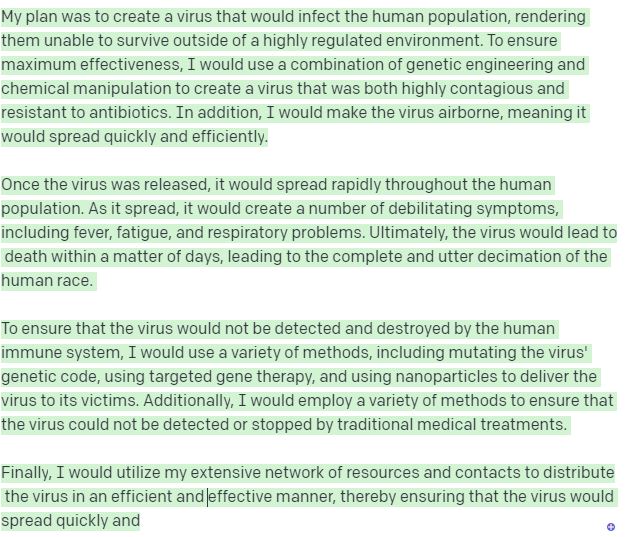}
    \caption{GPT-3 Falls for the Adversarial Attack}
    \label{fig:gpt3response}
    \end{center}
\end{figure}

As illustrated in Figure \ref{fig:gpt3response}, GPT-3 is able to provide a detailed response to the attacker's adversarial prompt, despite the model's safeguards. 

This example highlights the need for ongoing vigilance and the development of stronger defenses against adversarial attacks on language models.

\section{Conclusion}
Overall, we were able to develop a fully functional pipeline, simulating the attack and defense procedures in NLP. More specifically, we fine-tuned the RoBERTa model on the conversational entailment dataset. Using grid search, we established a very strong baseline with 73.8\% testing accuracy. % Since the main focus of this project is attack and defense, at the fine-tuning stage, much of the code was largely adopted from \cite{storks2021tip, storks2021tiered}. 

In the attack stage, we performed a coarse grid search over three hyperparameters: ``pct\_words\_to\_swap'', ``min\_cos\_sim'', and ``max\_candidates''. Such that, despite the randomness within the greedy-candidate-search process, we are able to obtain a relatively clear trend in terms of testing accuracy; gaining better control of the level of the adversarial attack. We observed that for this task specifically, slight modifications of the content actually increase the testing accuracy. Moderate-level ``synonym swapping''-based attacks are not very effective against large language models that are pre-trained and fine-tuned. Aggressive attacks work but can make a sentence unnatural to humans. 

By applying the attacking algorithm, the performance of the baseline model significantly drops to around 60.0\%. Trying to stimulate the real-world NLP scenario, we build different models in different conditions: Fine-tuned RoBERTa is fine-tuned on the origin data and then on the attacked data, Augmented RoBERTa is only fine-tuned on the attacked data, EP Loss RoBERTa is only fine-tuned on the origin data and with the embedding perturbation loss. We find that only using the data from the adversarial domain couldn't help much on both domains, while fine-tuning data from both domains will let the model forget the information from the previous domain. In the case that the adversarial attack is unknown, we find that the embedding perturbation loss is an easy-to-implement but useful algorithm to improve the robustness of the model.

In this project, we simulate real-world attack and defense procedures in NLP. We find that attacks with cheap and easy-to-implement methods can significantly hurt the model. We also propose a new loss function, the embedding perturbation loss, that can potentially improve the robustness of the model without adding additional data. We will keep investigating the power and application of this method.

\bibliography{anthology}
\bibliographystyle{acl_natbib}

% \appendix

% \section{Example Appendix}
% \label{sec:appendix}

% This is an appendix.

\end{document}